\documentclass[letterpaper]{article} 
\usepackage{aaai2026}  
\usepackage{times}  
\usepackage{helvet}  
\usepackage{courier}  
\usepackage[hyphens]{url}  
\usepackage{graphicx} 
\usepackage{natbib}  
\usepackage{caption} 
\usepackage{algorithm}
\usepackage{algorithmic}

\usepackage{booktabs} 
\usepackage{multirow} 
\usepackage{array}      
\usepackage{siunitx}    
\usepackage{graphicx}  
\usepackage{adjustbox}  
\usepackage{amsmath}    
\usepackage{enumitem}                
\usepackage[mathscr]{eucal}          
\usepackage{ragged2e}  
\usepackage{siunitx}   
\usepackage{tabularx}  
\usepackage{bm}  
\usepackage{soul}       
\usepackage{tabularx} 
\usepackage[table]{xcolor} 
\usepackage{amsfonts} 
\usepackage{amssymb}

\usepackage{newfloat}
\usepackage{listings}
\DeclareCaptionStyle{ruled}{labelfont=normalfont,labelsep=colon,strut=off} 
\floatstyle{ruled}
\newfloat{listing}{tb}{lst}{}
\floatname{listing}{Listing}
\title{PC-CrossDiff: Point-Cluster Dual-Level Cross-Modal Differential Attention for Unified 3D Referring and Segmentation}
\author{
	Wenbin Tan\textsuperscript{\rm 1,\rm 2},
	Jiawen Lin\textsuperscript{\rm 1}, 
	Fangyong Wang\textsuperscript{\rm 3},
	Yuan Xie\textsuperscript{\rm 4},
	Yong Xie\textsuperscript{\rm 5}\thanks{Corresponding Author.},
	Yachao Zhang\textsuperscript{\rm 1}\footnotemark[\value{footnote}], \\
	Yanyun Qu\textsuperscript{\rm 1}
}

\affiliations{
     \textsuperscript{\rm 1}Key Laboratory of Multimedia Trusted Perception and Efficient Computing, Ministry of Education of China, School of Informatics, Xiamen University, Xiamen, China\\
    \textsuperscript{\rm 2}School of Big Data, Tongren University, Tongren, China\\
     \textsuperscript{\rm 3}Hanjiang National Laboratory, Wuhan, China\\
     \textsuperscript{\rm 4}School of Computer Science and Technology, East China Normal University, Shanghai, China\\
     \textsuperscript{\rm 5}Department of Computer Science, Nanjing University of Posts and Telecommunications, Nanjing, China\\

     wbtan@stu.xmu.edu.cn, yongxie@njupt.edu.cn, yachaozhang@xmu.edu.cn
}

\usepackage{bibentry}

\begin{document}

\maketitle

\begin{abstract}
3D Visual Grounding (3DVG) aims to localize the referent of natural language referring expressions through two core tasks: Referring Expression Comprehension (3DREC) and Segmentation (3DRES). While existing methods achieve high accuracy in simple, single-object scenes, they suffer from severe performance degradation in complex, multi-object scenes that are common in real-world settings, hindering practical deployment. Existing methods face two key challenges in complex, multi-object scenes: inadequate parsing of implicit localization cues critical for disambiguating visually similar objects, and ineffective suppression of dynamic spatial interference from co-occurring objects, resulting in degraded grounding accuracy. To address these challenges, we propose PC-CrossDiff, a unified dual-task framework with a dual-level cross-modal differential attention architecture for 3DREC and 3DRES. Specifically, the framework introduces: (i) Point-Level Differential Attention (PLDA) modules that apply bidirectional differential attention between text and point clouds, adaptively extracting implicit localization cues via learnable weights to improve discriminative representation; (ii) Cluster-Level Differential Attention (CLDA) modules that establish a hierarchical attention mechanism to adaptively enhance localization-relevant spatial relationships while suppressing ambiguous or irrelevant spatial relations through a localization-aware differential attention block. To address the scale disparity and conflicting gradients in joint 3DREC–3DRES training, we propose $\mathcal{L}_{\text{DGTL}}$, a unified loss function that explicitly reduces multi-task crosstalk and enables effective parameter sharing across tasks. Our method achieves state-of-the-art performance on the ScanRefer, NR3D, and SR3D benchmarks. Notably, on the Implicit subsets of ScanRefer, it improves the Overall@0.50 score by $\textbf{+10.16\%}$ for the 3DREC task, highlighting its strong ability to parse implicit spatial cues.

\end{abstract}

\begin{links}
	\link{Code}{https://github.com/tanwb/PC-CrossDiff}.
\end{links}


\begin{figure}
	\centering
	\includegraphics[width=0.86\columnwidth]{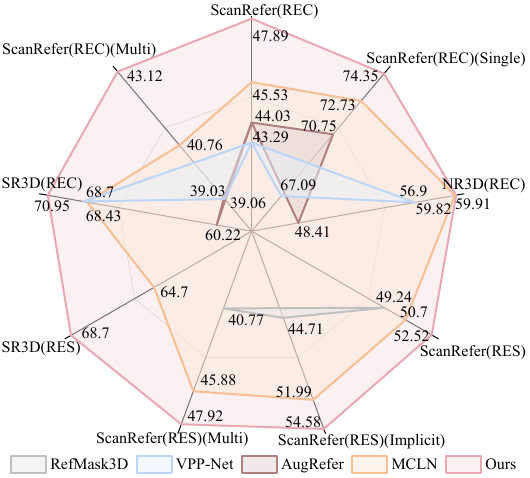}
	\caption{PC-CrossDiff vs. SOTA on 3DREC and 3DRES. ScanRefer (IoU@0.50), NR3D/SR3D (IoU@0.25).}
	\label{fig:motivation}
\end{figure}

\section{Introduction}

3D Visual Grounding (3DVG) enables machines to interpret natural language references in 3D scenes, with two core subtasks: 3D Referring Expression Comprehension (3DREC) for instance-level localization and 3D Referring Expression Segmentation (3DRES) for point-wise segmentation~\cite{zhangCrossModalMatchLanguage2024,xuMultiAttributeInteractionsMatter2024,wu3DSTMNDependencyDrivenSuperpointText2024}. These tasks are complementary, as 3DREC identifies target objects while 3DRES delineates their boundaries. For example, warehouse robots must detect dynamic obstacles via 3DREC and segment their boundaries via 3DRES. However, addressing these tasks independently fails to leverage shared cross-modal information to disambiguate multi-object scenes, necessitating a unified framework that balances computational efficiency with robust grounding accuracy, especially in complex multi-object scenes.

This need is further emphasized by the limitations of existing single-task approaches. As shown in Fig.~\ref{fig:motivation} on ScanRefer~\cite{chenScanRefer3DObject2020a}, state-of-the-art methods achieve high accuracy only in rare single-object settings but degrade significantly in multi-object scenes due to visual ambiguity. Specifically, these methods achieve Unique@0.50$>$70\% in single-object scenes ($15\%$ of cases), yet their performance drops below $46\%$ in multi-object scenes ($85\%$ of cases), revealing limited robustness in complex multi-object scenes.

This performance gap reveals two fundamental limitations. 
(i) Inadequate parsing of implicit localization cues (such as ``with magazines'' or ``beside the laptop''), which are crucial for disambiguating visually similar objects but remain underutilized due to the limited capacity of current frameworks. 
(ii) Insufficient suppression of dynamic spatial interference. In multi-object scenes, co-occurring objects introduce ambiguous spatial relations that existing methods aggregate indiscriminately, which amplifies noise and degrades performance.

Although extensive work has been conducted on single-task 3DREC and 3DRES, such as text--point cloud alignment~\cite{wuEDAExplicitTextDecoupling2023}, attribute integration~\cite{xuMultiAttributeInteractionsMatter2024}, and spatial modeling~\cite{wangG^3LQMarryingHyperbolic2024}, methods based on textual semantic parsing still primarily target explicit spatial relations (e.g., ``A--top--B'') and inadequately capture implicit cues that are critical for disambiguation in complex scenes. While graph-based models~\cite{fengFreeFormDescriptionGuided2021,huangTextGuidedGraphNeural2021} and relational CNNs~\cite{wangDynamicGraphCNN2019a} capture spatial dependencies, they lack mechanisms to dynamically suppress interference from irrelevant objects. Although MCLN~\cite{qianMultibranchCollaborativeLearning2024} advances dual-task co-learning, the development of a unified framework robust to multi-object interference remains a significant challenge.

To bridge this gap, we propose PC-CrossDiff: a Point-Cluster Cross-modal Differential framework for unified 3D referring and segmentation. The proposed framework addresses the core limitations through two key innovations:

(1) For implicit cue parsing, we design a lightweight Point-Level Differential Attention (PLDA) module. Unlike conventional methods requiring explicit entity/relation parsing and graph construction, PLDA processes full sentences via cross-modal differential attention between point clouds and text to extract visually grounded features, including implicit cues, through a learnable weighting mechanism that adaptively focuses on localization-relevant features while suppressing irrelevant ones, enabling target disambiguation in multi-object scenes. 

(2) For dynamic spatial filtering, we introduce Cluster-Level Differential Attention (CLDA). After modeling spatial relations among candidate points to capture global positional context and target-specific features, CLDA employs a Localization-Aware Differential Attention (LDA) block to selectively enhance localization-relevant spatial features while suppressing noise, enabling dynamic spatial relation filtering for robustness.

Furthermore, we propose the Dual-Geometry Task-Harmonized Loss ($\mathcal{L}_{\text{DGTL}}$) to resolve scale disparity and gradient conflicts in joint 3DREC–3DRES training, reducing multi-task crosstalk and enabling effective parameter sharing across tasks.

In summary, our contributions are: 
\begin{itemize} 
	
	\item We propose PC-CrossDiff, a unified cross-modal differential learning framework operating at both point and cluster levels. By synergistically integrating local perception with global spatial filtering, it enables competitive performance on both 3DREC and 3DRES within a single architecture.

	\item We design PLDA to extract implicit spatial cues directly from full-text descriptions via cross-modal differential attention, supporting accurate disambiguation in complex multi-object scenes; CLDA at the cluster level enables dynamic spatial relation filtering by selectively enhancing localization-relevant features through a localization-aware attention mechanism, improving robustness in complex multi-object scenes.

	\item Extensive experiments on ScanRefer, NR3D/SR3D, and their Implicit/Multiple subsets demonstrate that PC-CrossDiff achieves state-of-the-art performance on both 3DREC and 3DRES, significantly outperforming prior approaches.

\end{itemize}

\begin{figure*}
	\centering 
	\includegraphics[width=\linewidth]{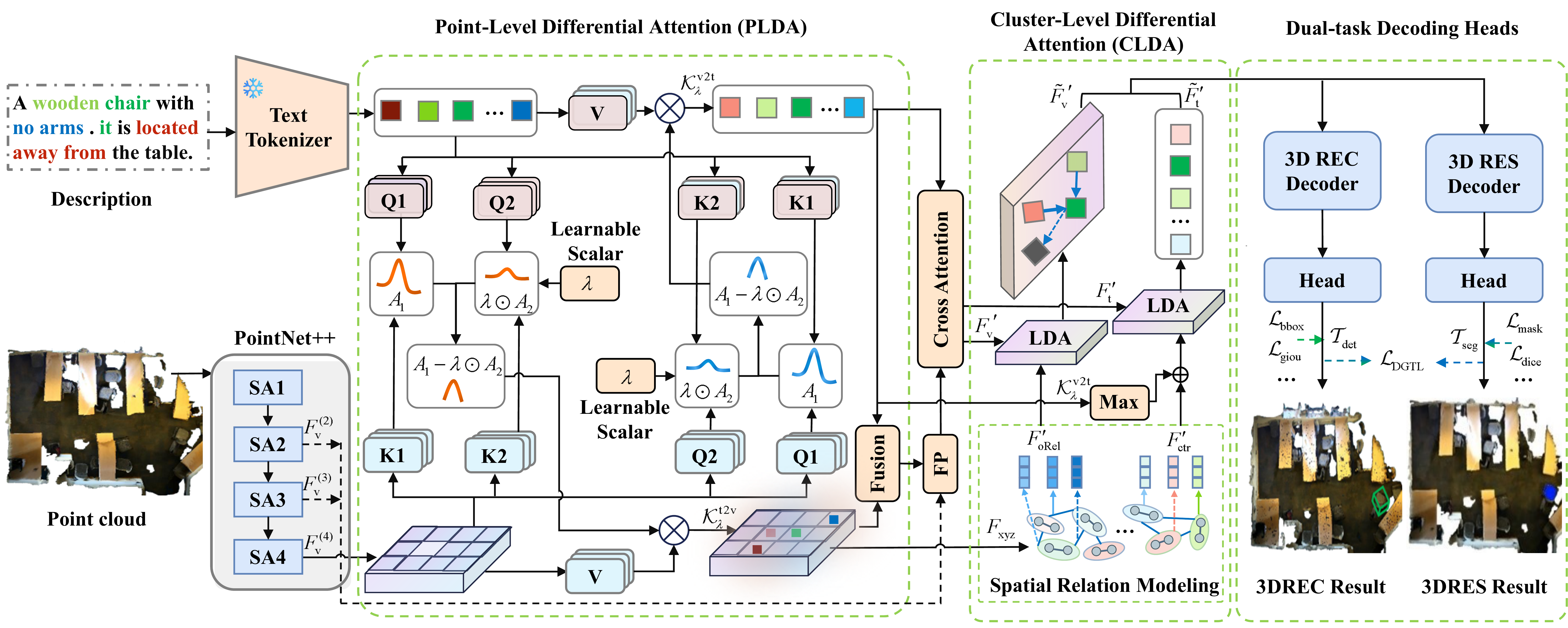} 
	\caption{The framework of PC-CrossDiff. Text and visual features are input into the PLDA module to extract implicit localization cues and enhance point-level visual features. Simultaneously, the coordinates $F_{\mathrm{xyz}}$ are fed into the CLDA module for Spatial Relation Modeling (SRM), modeling cluster-level relations. These features are enhanced by LDA and integrated into the aligned features $F'_{\text{v}}$ and $F'_{\text{t}}$, yielding $\tilde{F}_\text{v}^{\prime}$ and $\tilde{F}_\text{t}^{\prime}$. The enhanced features are decoded by 3DREC and 3DRES heads to generate predictions, with $\mathcal{L}_{\text{DGTL}}$ facilitating task synergy.}
	\label{fig_overall_network}
\end{figure*}

\section{Related work}
\subsection{3D Visual Grounding}

Advances in 3DVG are driven by the growing demand for indoor applications. Current methods can be categorized into two subtasks: 3DREC~\cite{zhangCrossModalMatchLanguage2024,xuMultiAttributeInteractionsMatter2024} and 3DRES~\cite{wu3DSTMNDependencyDrivenSuperpointText2024,qianXRefSeg3DEnhancingReferring2024}. 3DREC is typically implemented via two-stage architectures, which use 3D object detectors or RPNs for proposal generation before cross-modal alignment~\cite{zhangCrossModalMatchLanguage2024,achlioptasReferIt3DNeuralListeners2020}, or via single-stage approaches that directly establish visual-textual correspondences~\cite{luo3DSPSSingleStage3D2022,jainBottomTopDetection2022}. Research on 3DRES remains relatively limited.

Recent progress focuses on enhancing textual semantic understanding in 3D scenes~\cite{fengFreeFormDescriptionGuided2021,wuCLIP2UDAMakingFrozen2024} and on improving cross-modal feature correspondences~\cite{wuEDAExplicitTextDecoupling2023,qianXRefSeg3DEnhancingReferring2024,wuCrossmodalUnsupervisedDomain2023,wu2025fusion}. Transformer-based frameworks~\cite{jainBottomTopDetection2022,heTransRefer3DEntityandRelationAware2021,zhao3DVGTransformerRelationModeling2021,wuEDAExplicitTextDecoupling2023} excel through attention-driven visual-textual fusion, alongside approaches in point cloud segmentation~\cite{zhang2025cross,zhang2021perturbed}. Crucially, unified modeling of both tasks remains rare; MCLN~\cite{qianMultibranchCollaborativeLearning2024} is the only method that attempts joint modeling of 3DREC and 3DRES, yet it suffers severe performance degradation in multi-object scenes, limiting its practical deployment. This limitation motivates our design of a robust, unified framework.

\subsection{Spatial Positional Relationship Extraction}

In recent years, modeling spatial relationships in point clouds has become a focal research area for context-aware feature learning~\cite{wuUniDSegUnifiedCrossDomain2024}. Early CNN-based approaches~\cite{wangDynamicGraphCNN2019a,liuRelationShapeConvolutionalNeural2019} encode positional relations through local geometric feature extraction. Subsequent graph-based methods~\cite{wangOmniQOmniDirectionalScene2024,fengFreeFormDescriptionGuided2021} capture inter-object spatial dependencies. Recent advances further enhance spatial understanding; for example, TGNN~\cite{huangTextGuidedGraphNeural2021} improves local contextual modeling with graph neural networks, while MA2TransVG~\cite{xuMultiAttributeInteractionsMatter2024} and G$^{3}$LQ~\cite{wangG^3LQMarryingHyperbolic2024} leverage multi-attribute interactions and geometric perception, respectively.

However, existing methods lack dynamic filtering of spatial interference, making it difficult to distinguish hierarchical relationships in multi-object scenes. We address this limitation by introducing a cluster-level differential attention mechanism.

\section{Methods}

\subsection{Overview of PC-CrossDiff}
\label{sec:overview}

To enhance localization robustness in multi-object scenes, PC-CrossDiff integrates PLDA and CLDA into a unified differential learning framework. This two-level design synergistically combines local perception with global spatial filtering, enabling precise visual-text alignment and dynamic suppression of interference from co-occurring objects. As shown in Fig.~\ref{fig_overall_network}, PC-CrossDiff adopts a DETR-like architecture with dual decoders for joint 3DREC and 3DRES.

The input point cloud $\mathcal{P}$ and text $T$ are processed via PointNet++ \cite{qiPointNetDeepHierarchical2017a} and pretrained RoBERTa \cite{liuRoBERTaRobustlyOptimized2019} feature extractors. This yields textual features $F_{\text{t}} \in \mathbb{R}^{l_{\text{t}} \times d_{\text{emb}}}$ ($l_{\text{t}}$: tokens, $d_{\text{emb}}$: embedding dim) and high-level visual features $F_{\text{v}}^{(4)} \in \mathbb{R}^{n_{\mathrm{sem}} \times d_{\mathrm{sem}}}$ ($n_{\mathrm{sem}}$: points, $d_{\mathrm{sem}}$: semantic dim). The PLDA module computes fine-grained cross-modal localization features $F'_{\text{PLDA}} \in \mathbb{R}^{n_{\mathrm{sem}} \times d_{\mathrm{sem}}}$ incorporating implicit cues. These features are fused with intermediate PointNet++ features $F_{\text{v}}^{(3)}$ and $F_{\text{v}}^{(2)}$ via upsampling and aligned cross-modally to produce $F'_{\text{v}}$ and $F'_{\text{t}}$. To mitigate interference from multi-level spatial relationships in $F'_{\text{v}}$ within multi-object scenes, $F'_{\text{t}}$, $F'_{\text{v}}$, and positional coordinates $F_{\mathrm{xyz}} \in \mathbb{R}^{n_{\mathrm{p}} \times 3}$ ($n_{\mathrm{p}}$: points) are processed by the CLDA module to enhance localization-related features. The refined features feed into dual decoders.

\subsection{Point-Level Differential Attention}
\label{sec:PLDA}

Inspired by the Differential Transformer~\cite{yeDifferentialTransformer2024}, we propose the Point-Level Differential Attention (PLDA) module to extract implicit spatial cues from descriptions, which enhances discriminative features via a single-layer bidirectional cross-modal differential attention mechanism. The module comprises two symmetric components: 
(1) \textit{Text-to-Visual}: Enhances point-level visual features using textual localization references;  
(2) \textit{Visual-to-Text}: Extracts implicit localization cues from text guided by visual features.
\paragraph{Visual-to-Text Cross-modal Differential Attention.} This layer projects high-level visual features $F_{\mathrm{v}}^{(4)}$ to queries ($Q$) and textual features $T$ to keys ($K$)/values ($V$) via learnable matrices $W^Q, W^K, W^V \in \mathbb{R}^{d_\mathrm{sem} \times d_\mathrm{sem}}$:
\begin{equation}
	Q = F_{\mathrm{v}}^{(4)} W^Q, \quad 
	K = T W^K, \quad 
	V = T W^V.
	\label{eq:projection}
\end{equation}
The projections are reshaped into multi-head structures (head dimension $d_\mathrm{h} = d_\mathrm{sem}/(2N_\mathrm{h})$, $N_\mathrm{h}=8$):
\begin{equation}
	\begin{aligned}
		Q &= \text{reshape}\left(Q, [n_\mathrm{sem}, 2N_\mathrm{h}, d_\mathrm{h}]\right)^\intercal = [Q_1, Q_2], \\
		K &= \text{reshape}\left(K, [l_\mathrm{t}, 2N_\mathrm{h}, d_\mathrm{h}]\right)^\intercal = [K_1, K_2], \\
		V &= \text{reshape}\left(V, [l_\mathrm{t}, N_\mathrm{h}, 2d_\mathrm{h}]\right)^\intercal.
	\end{aligned}
	\label{eq:q_k_v}
\end{equation}
Then dual attention kernels are computed as:
\begin{equation}
	A_1 = \text{softmax}\left(\frac{Q_1 K_1^\intercal}{\sqrt{d_\mathrm{h}}}\right), 
	A_2 = \text{softmax}\left(\frac{Q_2 K_2^\intercal}{\sqrt{d_\mathrm{h}}}\right).
	\label{eq:a_1_a_2}
\end{equation}
A learnable parameter $\lambda$ suppresses irrelevant attention:
\[
\lambda = \exp\left(\sum (\lambda_{\mathbf{q}_1} \odot \lambda_{\mathbf{k}_1})\right) - \exp\left(\sum (\lambda_{\mathbf{q}_2} \odot \lambda_{\mathbf{k}_2})\right),
\]
where $\lambda_{\mathbf{q}_1}, \lambda_{\mathbf{k}_1}, \lambda_{\mathbf{q}_2}, \lambda_{\mathbf{k}_2} \in \mathbb{R}^{d_\mathrm{h}}$ are learnable vectors. The visual-to-text cross-modal differential attention is then computed as:
\begin{equation}
	\text{DiffAttn}(F_{\mathrm{v}}^{(4)}, T) = (A_1 - \lambda \odot A_2)V.
	\label{eq:DiffAttn_fus_v}
\end{equation}
After layer normalization and linear projection, we obtain the visual-to-text cross-modal semantic feature $\mathcal{K}_\mathrm{v2t}$, encoding implicit localization cues for subsequent fusion.

\paragraph{Text-to-Visual Cross-modal Differential Attention.} This component enhances point-level visual features by using textual features $T$ as queries ($Q$) and $F_{\mathrm{v}}^{(4)}$ as keys ($K$)/values ($V$). The differential attention operation (Eqs.~\ref{eq:projection}--\ref{eq:DiffAttn_fus_v}) is applied with the query and key roles interchanged. The learnable $\lambda$ suppresses irrelevant visual features guided by the textual features, yielding enhanced point-level visual features $\mathcal{K}_\mathrm{t2v}$. For more details, see the Appendix.

\subsection{Cluster-Level Differential Attention}
\label{sec:CLDA}

\begin{figure}[t]
	\centering
	\includegraphics[width=\linewidth]{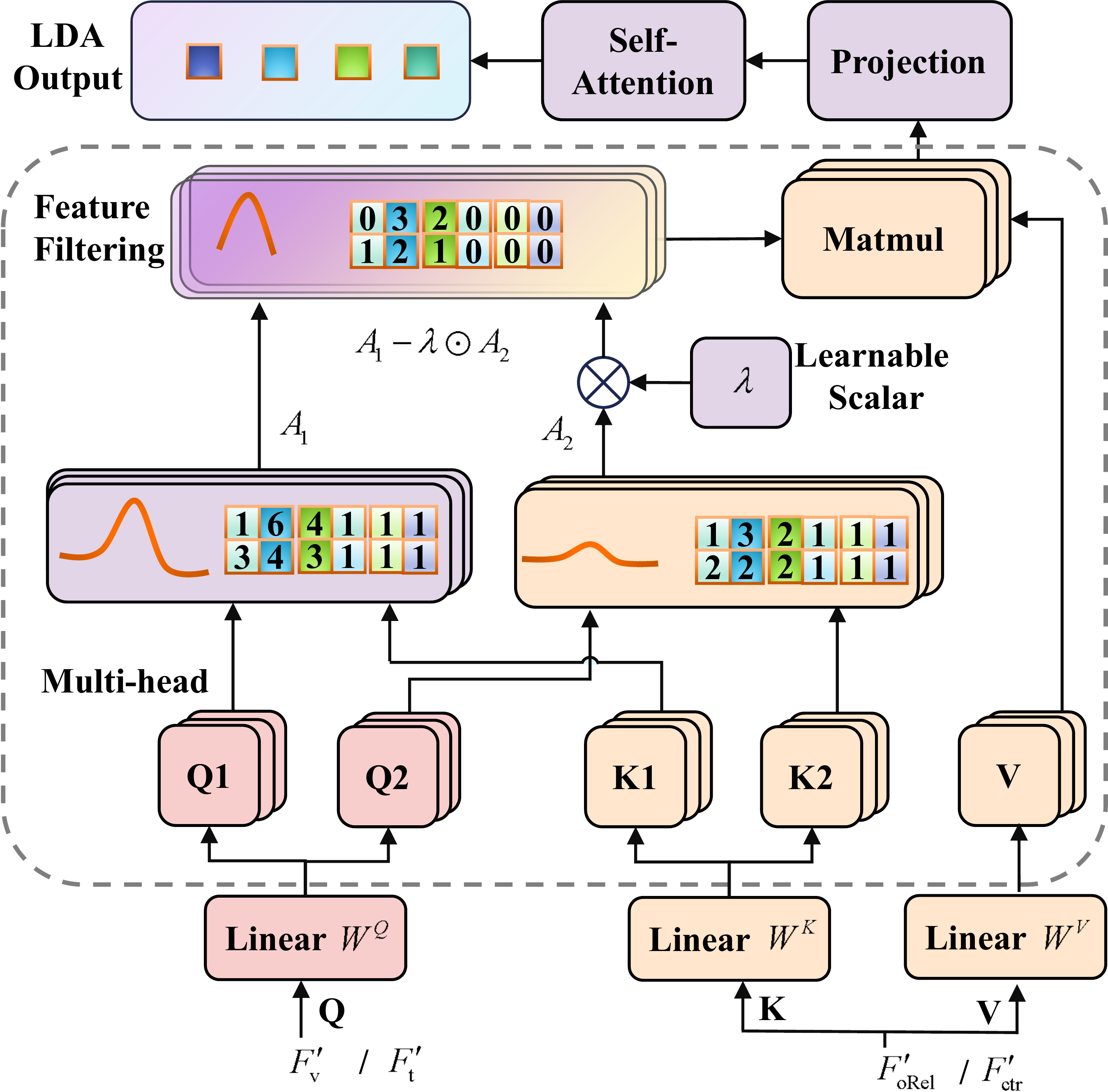}	
	\caption{LDA block. With $\mathit{F'}_{\mathrm{v}}$ as query, DiffAtten (\text{Eq.~\ref{eq:f_v_oRel}}) extracts relevant relations from $\mathit{F'}_{\mathrm{oRel}}$ while suppressing irrelevant ones. Features then pass self-attention to yield $\mathit{\tilde{F}}_{\mathrm{v}}^{\prime}$.}	
	\label{fig_lda}
\end{figure}

To address the challenge of interference from irrelevant objects in multi-object scenes, we propose the Cluster-Level Differential Attention (CLDA) module for dynamic spatial filtering. CLDA first models cluster-level spatial relations with the Spatial Relation Modeling (SRM) block and subsequently applies two Localization-Aware Differential Attention (LDA) blocks to selectively enhance target-relevant features while suppressing noise, thereby achieving robust, dynamic spatial filtering in the presence of multi-object interference (Fig.~\ref{fig_lda}).

\paragraph{Spatial Relation Modeling (SRM).} 
Given position coordinates $F_\text{xyz} \in \mathbb{R}^{n_\mathrm{p} \times 3}$ from $F'_{\mathrm{v}}$ (where $F_{\mathrm{v}}$ and $F'_{\mathrm{v}}$ share identical coordinates), SRM applies farthest point sampling (FPS) \cite{qiPointNetDeepHierarchical2017} to select $N_{\mathrm{clust}}$ candidate centroids. KNN clustering partitions points into $N_{\mathrm{clust}}$ clusters of size $S_\mathrm{clust}=n_\mathrm{p}/N_\mathrm{clust}$ ($n_\mathrm{p}=1024$).

An encoder processes intra-cluster points to obtain relative spatial relationships $F_\mathrm{iRel} \in \mathbb{R}^{N_{\mathrm{clust}} \times d_\mathrm{mod}}$. Inter-cluster spatial relationships are extracted via DGCNN \cite{wangDynamicGraphCNN2019a}:
\begin{equation}
	F'_{\text{oRel}} = \text{DGCNN}(F_{\mathrm{iRel}}, \mathbf{O}) \in \mathbb{R}^{N_{\text{clust}} \times d_{\text{mod}}},
\end{equation}
where $\mathbf{O} \in \mathbb{R}^{N_{\text{clust}} \times 3}$ represents cluster centroids, and candidate target region features are computed using a linear projection 
$F'_{\text{ctr}} = \mathbf{O} \mathbf{W}_\mathrm{o} + \mathbf{b}_\mathrm{o}$, 
where $\mathbf{W}_\mathrm{o} \in \mathbb{R}^{3 \times d_{\text{mod}}}$.

\paragraph{Localization-Aware Differential Attention (LDA).} 
LDA implements a two-stage process: (1) feature filtering via single-layer unidirectional differential attention, and (2) feature enhancement via multi-head self-attention \cite{yuPointBERTPreTraining3D2022}, as shown in Fig.~\ref{fig_lda}.

For spatial relationship filtering, aligned visual features $F'_{\mathrm{v}}$ serve as queries ($Q$), while inter-cluster relationships $F'_{\mathrm{oRel}}$ act as keys ($K$) and values ($V$). The computation follows Eqs.~\ref{eq:projection}--\ref{eq:DiffAttn_fus_v}:
\begin{equation}
	\text{DiffAttn}(F'_{\mathrm{v}}, F'_{\mathrm{oRel}}) = (A_1 - \lambda \odot A_2)V.
	\label{eq:f_v_oRel}
\end{equation}
This extracts and enhances localization-relevant features from $F'_{\mathrm{oRel}}$ into $F'_{\mathrm{v}}$, followed by self-attention refinement, yielding spatially enhanced features $\tilde{F}_{\mathrm{v}}^{\prime}$.
Similarly, using $F'_{\mathrm{t}}$ as queries ($Q$) and candidate regions $F'_{\mathrm{ctr}}$ as keys ($K$) and values ($V$), the computation is performed via Eqs.~\ref{eq:projection}--\ref{eq:DiffAttn_fus_v}. The computation selects localization-relevant target region features from $F'_{\mathrm{ctr}}$ and enhances $F'_{\mathrm{t}}$ with them, refining target region features via self-attention. See Appendix for details.

\subsection{Training Loss}

We propose the Dual-Geometry Task-Harmonized Loss ($\mathcal{L}_{\text{DGTL}}$) for joint 3DREC and 3DRES optimization:

\paragraph{Dual-Geometry Consistency.} 
We enforce bidirectional alignment between localization ($\mathbf{B}_{\text{loc}}$) and segmentation ($\mathbf{M}$) geometries to establish structural coherence:
\[
\mathcal{L}_{\mathrm{geom}} = \zeta(t) \left[\mathcal{L}_{\mathrm{IoU}}(\mathbf{B}_{\mathrm{loc}}, \mathbf{B}_{\text{mask}}) + \mathcal{L}_{\mathrm{Dice}}(\mathbf{M}, \mathbf{M}_{\text{box}})\right],
\]
where $\zeta(t)$ implements a linear warm-up, $\mathcal{L}_{\mathrm{IoU}}$ and $\mathcal{L}_{\mathrm{Dice}}$ are IoU and Dice losses, and $\mathbf{B}_{\text{mask}}$ (mask-derived box) with $\mathbf{M}_{\text{box}}$ (box-derived mask) form a self-supervised consistency loop via internal predictions.

\paragraph{Contribution Balancing.}
We balance contributions via $w_i = \max(e^{-v_i}, \lambda_i)$, where $v_i$ is learnable and $\lambda_i$ is an empirical threshold from MCLN~\cite{qianMultibranchCollaborativeLearning2024}, ensuring meaningful task ratios.

\paragraph{Cross-task Harmonization.}
To harmonize optimization directions between detection ($\mathcal{T}_{\text{det}}$) and segmentation ($\mathcal{T}_{\text{seg}}$), we penalize conflicting gradients as:
$$
\mathcal{P} = \eta(t) \cdot \sum_{i \in \mathcal{T}_{\text{det}},\ j \in \mathcal{T}_{\text{seg}}} \rho_{ij} \cdot \mathbb{I}\left[\cos(\theta_{ij}) < \tau\right] \cdot \left(\tau - \cos(\theta_{ij})\right),
$$
where $\cos(\theta_{ij})$ measures gradient alignment, $\eta(t)$ is a linear decay factor, $\tau$ represents a threshold, $\mathbb{I}[\cdot]$ is the indicator function, and $\rho_{ij}$ denotes task correlations. 

The full loss $\mathcal{L}_{\text{DGTL}}$ (denoted as $\mathcal{L}$) integrates all components:
\begin{equation}
	\mathcal{L} = 
	\underbrace{
		\sum_{i \in \mathcal{T}_{\text{det}}} w_i\mathcal{L}_i + 
		\sum_{j \in \mathcal{T}_{\text{seg}}} w_j\mathcal{L}_j + 
		\mathcal{P}
	}_{\text{Task-Harmonized}} 
	+ \underbrace{
		\mathcal{L}_{\text{geom}}
	}_{\text{Dual-Geometry}},
	\label{l_dgtl}
\end{equation}
with dynamically learned weights $w_i$ and $\mathcal{L}_{\text{geom}}$ enforcing spatial consistency.

\section{Experiments}

\subsection{Experimental Setup}

We evaluate PC-CrossDiff against the state-of-the-art multi-task baseline MCLN~\cite{qianMultibranchCollaborativeLearning2024} on ScanRefer~\cite{chenScanRefer3DObject2020a}, NR3D/SR3D~\cite{achlioptasReferIt3DNeuralListeners2020}, and their challenging subsets (Implicit and Multiple), under identical protocols for fair comparison across 3DREC and 3DRES tasks. Segmentation annotations strictly follow ScanNet's official standards~\cite{daiScanNetRichlyAnnotated3D2017}. All experiments run on NVIDIA 3090 GPUs. We employ AdamW optimizer with learning rates of $2\times10^{-3}$ for visual encoders and $2\times10^{-4}$ for other layers. The temperature coefficient $\tau = 0.5$ and the number of network layers $L = 6$. Evaluation metrics and remaining hyperparameters align with MCLN's standards.

\subsection{Quantitative Comparisons}

\paragraph{Performance on ScanRefer (3DRES).} We conduct comparative experiments on the ScanRefer dataset with TGNN\cite{huangTextGuidedGraphNeural2021} and current state-of-the-art methods: X-RefSeg3D\cite{qianXRefSeg3DEnhancingReferring2024}, 3D-STMN\cite{wu3DSTMNDependencyDrivenSuperpointText2024}, SegPoint\cite{heSegPointSegmentAny2025}, MCLN\cite{qianMultibranchCollaborativeLearning2024}, and RefMask3D\cite{heRefMask3DLanguageGuidedTransformer2024}. As shown in Table~\ref{scanrefer_3dres}, our PC-CrossDiff achieves superior performance, achieving 60.41\%, 52.52\%, and 46.39\% on Overall@0.25, Overall@0.50, and mIoU metrics, respectively. Specifically, PC-CrossDiff achieves improvements of 20.08\%, 5.81\%, and 4.54\% over X-RefSeg3D, 3D-STMN, and RefMask3D on Overall@0.25. Similarly, it improves by 18.75\%, 12.72\%, and 3.28\% on Overall@0.50. These results demonstrate that PC-CrossDiff achieves significant performance advantages in 3DRES tasks while maintaining a high inference speed of 4.07 FPS. This is attributed to the lightweight PLDA and CLDA modules, which enhance the model's robustness in multi-object environments via improved text understanding (especially implicit localization cues) and spatial filtering.

\paragraph{Performance on Multiple and Implicit Subsets.}
To evaluate PC-CrossDiff's performance in multi-object scenes, we conduct experiments on the Multiple subset of ScanRefer. To further assess its capacity to capture implicit localization cues, we construct an Implicit subset comprising 501 referring expressions with implicit references from the ScanRefer validation set (see Appendix). We use the model weights trained under the settings of Table~\ref{scanrefer_3dres} for inference, and the results are summarized in Table~\ref{implicit_res}.
On the Multiple@0.50 metric, PC-CrossDiff achieves improvements of 24.31\%, 18.72\%, and 7.15\% over X-RefSeg3D~\cite{qianXRefSeg3DEnhancingReferring2024}, 3D-STMN~\cite{wu3DSTMNDependencyDrivenSuperpointText2024}, and RefMask3D~\cite{heRefMask3DLanguageGuidedTransformer2024}, respectively.

On the Implicit subset, our method obtains a 17.25\% gain over X-RefSeg3D and a 14.26\% improvement over 3D-STMN in Implicit@0.50. Notably, it outperforms MCLN (baseline) by +10.16\% in 3DREC. These results demonstrate PC-CrossDiff’s superior ability to handle complex multi-object scenes and effectively exploit implicit spatial cues.

\begin{table}
	\centering
	
	\begin{tabular}{@{\extracolsep{\fill}} 
			>{\centering\arraybackslash}p{0.23\linewidth}
			>{\centering\arraybackslash}p{0.2\linewidth}
			*{3}{>{\centering\arraybackslash}p{0.07\linewidth}} 
			>{\centering\arraybackslash}p{0.07\linewidth}
		}
		\toprule
		\textbf{Method} &   \textbf{Venue}  & \multicolumn{2}{c}{\textbf{Overall}} & \textbf{mIoU}& \textbf{FPS} \\
		\cmidrule(lr){3-4} 
		&  & \textbf{0.25} & \textbf{0.50} & & \\
		\midrule

		TGNN &   AAAI'21 & 37.50 & 31.40 & 27.80 & 2.85 \\
		X-RefSeg3D & AAAI'24 & 40.33 & 33.77 & 29.94 & 2.03 \\
		
		3D-STMN & AAAI'24 & 54.60 & 39.80 & 39.50 & 4.14 \\

		SegPoint & ECCV'24 & - & - & 41.70 & - \\

		MCLN & ECCV'24 & 58.70 & 50.70 & 44.72 & 4.81\\

		RefMask3D & ACMMM'24 & 55.87 & 49.24 & 44.86 & 1.65  \\

		Ours & / & \textbf{60.41} & \textbf{52.52} & \textbf{46.39} & 4.07 \\

		\bottomrule
		
	\end{tabular}
	\caption{Comparison results of 3DRES on the ScanRefer dataset (inference: batchsize=1, on an NVIDIA 3090 GPU).}
	\label{scanrefer_3dres}
\end{table}

\begin{table}
	\centering
	
	\begin{tabular}{@{\extracolsep{\fill}} 
			>{\centering\arraybackslash}p{0.24\linewidth}
			*{3}{>{\centering\arraybackslash}p{0.13\linewidth}} 
			>{\centering\arraybackslash}p{0.12\linewidth}
		}
		\toprule
		\textbf{Method} &  \multicolumn{2}{c}{\textbf{Multiple}} & \multicolumn{2}{c}{\textbf{Implicit}} \\
		\cmidrule(lr){2-3} 	\cmidrule(lr){4-5} 
		&   \textbf{0.25} & \textbf{0.50} & \textbf{0.25} & \textbf{0.50} \\
		\midrule
		\multicolumn{5}{l}{\cellcolor{gray!10}\textit{3DRES}} \\
		
		X-RefSeg3D & 28.16 & 23.61 & 43.71 & 37.33 \\
		
		3D-STMN &  46.20 & 29.20 & 57.07 & 40.32 \\

		RefMask3D &  48.09 & 40.77  & 49.30 & 44.71 \\
		
		MCLN &  53.28 & 45.88 & 59.16 & 51.99 \\
		
		Ours  & \textbf{55.33} & \textbf{47.92} & \textbf{62.15} & \textbf{54.58} \\
		\midrule	
		\rowcolor{gray!10}
		\multicolumn{5}{l}{\textit{3DREC}} \\
		
		MCLN &  51.96 & 40.76 & 54.18 & 38.45 \\
		
		Ours &  \textbf{53.59} & \textbf{43.12} & \textbf{60.76} & \textbf{48.61} \\
		\bottomrule
	\end{tabular}
	\caption{Joint Evaluation of 3DRES and 3DREC across Multiple and Implicit Subsets in the ScanRefer Dataset.}  	
	\label{implicit_res}
\end{table}

\paragraph{Inference Speed.}

PC-CrossDiff was evaluated under identical hardware and input settings. As shown in Table~\ref{scanrefer_3dres}, our method achieves 4.07~FPS, outperforming X-RefSeg3D~\cite{qianXRefSeg3DEnhancingReferring2024} and RefMask3D~\cite{heRefMask3DLanguageGuidedTransformer2024} in terms of both localization accuracy and inference speed. This efficiency stems from the lightweight design of the PLDA and CLDA modules; for instance, PLDA uses a single layer of cross-modal differential attention with only 2.10M~parameters, introducing a negligible increase in model size and complexity. See the Appendix for details.

\paragraph{Performance on ScanRefer (3DREC).}
Table~\ref{scanrefer_3drec} compares PC-CrossDiff with recent single-stage and two-stage methods. In the two-stage setting, we compare with MVT~\cite{huangMultiViewTransformer3D2022}, 3D-SPS~\cite{luo3DSPSSingleStage3D2022}, BUTD-DETR~\cite{jainBottomTopDetection2022}, ViL3DRel~\cite{chenLanguageConditionedSpatial2022}, 3D-VLP~\cite{jinContextawareAlignmentMutual2023}, EDA~\cite{wuEDAExplicitTextDecoupling2023}, 3D-VisTA~\cite{zhu3DVisTAPretrainedTransformer2023a}, VPP-Net~\cite{shiViewpointAwareVisualGrounding2024}, G3-LQ~\cite{wangG^3LQMarryingHyperbolic2024}, MA2TransVG~\cite{xuMultiAttributeInteractionsMatter2024}, and MCLN. PC-CrossDiff achieves 58.47\% and 47.89\% on Overall@0.25 and Overall@0.50, respectively, surpassing all baselines. On Multiple@0.50, it improves by 2.36\% over MCLN. In the single-stage setting, we compare with AugRefer~\cite{wangAugReferAdvancing3D2025a}, TSP3D~\cite{guoTextguidedSparseVoxel2025}, and other state-of-the-art models. PC-CrossDiff exceeds MCLN and AugRefer by 3.61\% and 2.96\% on Multiple@0.50, and improves by 4.04\% and 2.65\% on Overall@0.50. Notably, its Overall@0.50 score surpasses several two-stage models, including VPP-Net, G3-LQ, and MCLN.

\begin{table}
	\centering

	\begin{tabular}{ @{} 
			>{\centering\arraybackslash}p{0.24\linewidth}
			>{\centering\arraybackslash}p{0.14\linewidth}
			*{3}{>{\centering\arraybackslash}p{0.08\linewidth}} 
			>{\centering\arraybackslash}p{0.09\linewidth}  @{} 
		}
		\toprule
		\textbf{Method} & \textbf{Venue} &  \multicolumn{2}{c}{\textbf{Multiple}} & \multicolumn{2}{c}{\textbf{Overall}} \\
		\cmidrule(lr){3-4} \cmidrule(lr){5-6} 
		&  & \textbf{0.25} & \textbf{0.50} & \textbf{0.25} & \textbf{0.50} \\
		\midrule
		
		\rowcolor{gray!10}
		\multicolumn{6}{@{}l@{}}{\textit{Two-stage Model}} \\
		
		\midrule		
		\rowcolor{gray!10}	MVT & CVPR'22 & 31.92 & 25.26 & 40.80 & 33.26 \\
		\rowcolor{gray!10}	3D-SPS & CVPR'22 & 40.32 & 29.82 & 48.82 & 36.98 \\
		\rowcolor{gray!10}	BUTD-DETR & ECCV'22 & 44.73 & 33.97 & 50.42 & 38.60 \\
		\rowcolor{gray!10}	ViL3DRel & NIPS'22 & 40.30 & 30.71 & 47.94 & 37.73 \\
		\rowcolor{gray!10}	3D-VLP & CVPR'23 & 43.51 & 33.41 & 51.41 & 39.46 \\
		\rowcolor{gray!10}	EDA & CVPR'23 & 49.13 & 37.64 & 54.59 & 42.26 \\
		\rowcolor{gray!10}	3DRefTR-SP & ICCV'23 & 50.07 & 38.65 & 55.45 & 43.48 \\
		\rowcolor{gray!10}	3D-VisTA & ICCV'23 & 43.70 & 39.10 & 50.60 & 45.80 \\
		\rowcolor{gray!10}	VPP-Net & CVPR'24 & 50.53 & 39.03 & 55.65 & 43.29 \\
		\rowcolor{gray!10}	G3-LQ & CVPR'24 & 51.48 & 40.80 & 56.90 & 45.58 \\	
		\midrule
		
		\rowcolor{gray!10}
		\multicolumn{6}{@{}l@{}}{\textit{Dual-Task}} \\

		\rowcolor{gray!10}	MCLN & ECCV'24 & 51.96 & 40.76 & 57.17 & 45.53 \\

		\rowcolor{gray!10}	Ours & / & \textbf{53.59} & \textbf{43.12} & \textbf{58.47} & \textbf{47.89} \\

		\midrule
		\multicolumn{6}{l}{\textit{Single-stage Model}} \\
		\midrule
		3D-SPS & CVPR'22 & 39.48 & 29.61 & 47.65 & 36.43 \\
		BUTD-DETR & ECCV'22 & 44.20 & 32.81 & 49.76 & 37.05 \\
		EDA & CVPR'23 & 48.11 & 36.82 & 53.83 & 41.70 \\
		
		G3-LQ & CVPR'24 & 50.23 & 39.72 & 55.95 & 44.72 \\

		AugRefer & AAAI'25 & 49.96 & 39.06 & 55.68 & 44.03 \\
		TSP3D & CVPR'25   &   -    &    -   &    56.45   &  46.71      \\
		TSP3D* & CVPR'25   &  50.66   &    41.30   &    55.82   &    45.69    \\
		\midrule
		\multicolumn{6}{l}{\textit{Dual-task}} \\	
		MCLN  & ECCV'24 & 49.72 & 38.41 & 54.30 & 42.64 \\	
		Ours & / & \textbf{51.59} & \textbf{42.02} & \textbf{57.37} & \textbf{46.68} \\
		
		\bottomrule
	\end{tabular}
	\caption{Comparison results of single-stage and two-stage 3DREC on the ScanRefer dataset (*: Reproduction).}
	\label{scanrefer_3drec}
\end{table}

\begin{table}[htbp]
	\centering
	
	\begin{tabular}{
			>{\centering\arraybackslash}p{0.32\linewidth}
			>{\centering\arraybackslash}p{0.19\linewidth}
			>{\centering\arraybackslash}p{0.14\linewidth}
			>{\centering\arraybackslash}p{0.14\linewidth}
		}
		\toprule		
		\textbf{Method} & \textbf{Venue} & \textbf{SR3D} & \textbf{NR3D} \\
		\midrule
		
		3D-SPS & CVPR'22 & 62.60 & 51.50 \\
		MVT & CVPR'22 & 64.50 & 55.10 \\
		BUTD-DETR & ECCV'22 & 65.60 & 49.10 \\
		LAR & NIPS'22 & 59.60 & 48.90 \\
		EDA & CVPR'23 & 68.10 & 52.10 \\
		VPP-Net & CVPR'24 & 68.70 & 56.90 \\
		
		AugRefer & AAAI'25 & 60.22 & 48.41 \\
		
		\midrule
		\multicolumn{4}{l}{\cellcolor{gray!10}\textit{Dual-Task}} \\	
		MCLN & ECCV'24 & 68.43 & 59.82 \\ 
		Ours & / & \textbf{70.95} & \textbf{59.91} \\
		
		\bottomrule
	\end{tabular}
	\caption{3DREC performance on SR3D and NR3D (Acc@0.25IoU).}
	\label{SR3D_NR3D}
\end{table}

\paragraph{Performance on SR3D/NR3D.} Following MCLN's evaluation protocol, we compare PC-CrossDiff with 3D-SPS~\cite{luo3DSPSSingleStage3D2022}, MVT~\cite{huangMultiViewTransformer3D2022}, BUTD-DETR~\cite{jainBottomTopDetection2022}, LAR~\cite{bakrLookRefer2d2022}, EDA~\cite{wuEDAExplicitTextDecoupling2023}, VPP-Net~\cite{shiViewpointAwareVisualGrounding2024}, and MCLN. As shown in Table~\ref{SR3D_NR3D}, PC-CrossDiff achieves state-of-the-art performance on both benchmarks. Specifically, on SR3D's 3DREC task, PC-CrossDiff outperforms VPP-Net and MCLN by 2.25\% and 2.52\%, respectively, with a 2.71\% improvement over VPP-Net on NR3D, demonstrating strong generalization capability.

\begin{table}[htbp]
	\centering

	\begin{tabular}{@{\extracolsep{\fill}}
			>{\centering\arraybackslash}p{0.08\linewidth}
			>{\centering\arraybackslash}p{0.08\linewidth}
			>{\centering\arraybackslash}p{0.08\linewidth}
			*{2}{>{\centering\arraybackslash}p{0.07\linewidth}}
			*{3}{>{\centering\arraybackslash}p{0.07\linewidth}}
		}
		\toprule
		\multicolumn{3}{c}{\textbf{Component}} &
		\multicolumn{2}{c}{\textbf{3DREC}} &
		\multicolumn{3}{c}{\textbf{3DRES}} \\
		\cmidrule(r){1-3} \cmidrule(lr){4-5} \cmidrule(l){6-8}
		$\bm{\mathcal{L_\text{DGTL}}}$ & \textbf{PLDA} & \textbf{CLDA} &
		\textbf{0.25} & \textbf{0.50} &
		\textbf{0.25} & \textbf{0.50} & \textbf{mIoU} \\
		
		\midrule
		w/o & w/o & w/o & 56.76 & 44.86 & 58.43 & 49.66 & 43.85 \\
		$\checkmark$ & w/o & w/o & 56.95 & 45.50 & 58.49 & 50.86 & 44.78 \\
		$\checkmark$ & $\checkmark$ & w/o & 57.56 & 46.20 & 59.19 & 50.99 & 45.12 \\
		$\checkmark$ & w/o & $\checkmark$ & 57.69 & 46.18 & 59.61 & 51.88 & 45.68 \\
		$\checkmark$ & $\checkmark$ & $\checkmark$ & \textbf{58.47} & \textbf{47.89} & \textbf{60.41} & \textbf{52.52} & \textbf{46.39} \\

		\bottomrule
	\end{tabular}
	
	\caption{Ablation study of $\mathcal{L_\text{DGTL}}$, PLDA, and CLDA on 3DREC/3DRES ($\checkmark$: enabled; w/o: disabled).}
	
	\label{as_bcad_CLDA}
\end{table}

\subsection{Ablation Studies}

All ablation studies are conducted on the ScanRefer dataset to evaluate performance on both 3DREC and 3DRES tasks.

To systematically evaluate the $\mathcal{L_\text{DGTL}}$, PLDA, and CLDA modules, we conduct ablation studies on the ScanRefer benchmark. As shown in Table~\ref{as_bcad_CLDA}, introducing $\mathcal{L_\text{DGTL}}$ alone improves the Overall@0.50 score by 1.2\% on the 3DRES task, validating the necessity of the loss in mitigating learning imbalance. Adding PLDA alone boosts performance by 1.33\%, validating its ability to extract implicit localization cues through bidirectional attention and refine point-level features. Similarly, incorporating CLDA alone yields gains of 2.22\%, demonstrating its effectiveness in dynamic spatial relation modeling and noise suppression for robust localization in multi-object scenes. When all three components are combined, the model achieves optimal performance with notable improvements on the Multiple and Implicit subsets (Appendix 3.1), demonstrating synergy across components.

To evaluate the robustness of localization accuracy to cluster size, we analyze its impact. As shown in Table~\ref{as_group_num}, when the cluster size increases from $S_{\mathrm{clust}} = 8$ to $S_{\mathrm{clust}} = 16$, performance varies by less than 0.32\% for 3DREC@0.25. This stability indicates that our functional decoupling design effectively mitigates clustering-induced errors, with additional details provided in the Appendix.

\begin{table}
	\centering
	
	\begin{tabular}{@{\extracolsep{\fill}}
			>{\centering\arraybackslash}p{0.1\linewidth}
			>{\centering\arraybackslash}p{0.1\linewidth}
			*{2}{>{\centering\arraybackslash}p{0.09\linewidth}}
			*{3}{>{\centering\arraybackslash}p{0.09\linewidth}}
		}
		\toprule 
		\multicolumn{2}{c}{\textbf{CLDA}} & 
		\multicolumn{2}{c}{\textbf{3DREC}} & 
		\multicolumn{3}{c}{\textbf{3DRES}} \\
		\cmidrule(r){1-2} \cmidrule(lr){3-4} \cmidrule(l){5-7}
		\textbf{$N_{\mathrm{clust}}$} & 
		\textbf{$S_{\mathrm{clust}}$} & 
		\textbf{0.25} & \textbf{0.50} & 
		\textbf{0.25} & \textbf{0.50} & 
		\textbf{mIoU} \\
		\midrule
		256 & 4 & 56.11 & 46.37 & 57.83 & 50.88 & 44.66 \\
		128 & 8 & 58.15 & 47.09 & 60.21 & 52.40 & 46.16 \\
		64 & 16 & \textbf{58.47} & \textbf{47.89} & \textbf{60.41} & \textbf{52.52} & \textbf{46.39} \\
		32 & 32 & 57.36 & 47.01 & 59.25 & 51.53 & 45.54 \\
		\bottomrule
	\end{tabular}
	
	\caption{Ablation study of cluster configurations ($N_{\mathrm{clust}}$ and $S_{\mathrm{clust}}$) on 3DREC and 3DRES tasks.}
	\label{as_group_num}
	
\end{table}

We evaluate feature transfer from the PLDA module to the CLDA module's Max block (Fig.~\ref{fig_overall_network}) and analyze differential attention (DiffAtten, Eq.~\ref{eq:DiffAttn_fus_v}) through two experiments: 
(i) Comparing original visual features $F^{(4)}_{\mathrm{v}}$ with $\mathcal{K}^{\mathrm{t}2\mathrm{v}}_{\lambda}$ obtained via text-guided DiffAtten; 
(ii) Comparing original text features $F_{\mathrm{t}}$ with $\mathcal{K}^{\mathrm{v}2\mathrm{t}}_{\lambda}$ obtained via vision-guided DiffAtten.

\begin{table}[htbp]
	\centering
	\begin{tabular}{@{\extracolsep{\fill}} 
			>{\centering\arraybackslash}p{0.2\linewidth}
			>{\centering\arraybackslash}p{0.1\linewidth}
			*{3}{>{\centering\arraybackslash}p{0.1\linewidth}} 
			>{\centering\arraybackslash}p{0.1\linewidth}
		}
		\toprule 
		\textbf{Component} &  
		\multicolumn{2}{c}{\textbf{3DREC}} & 
		\multicolumn{3}{c}{\textbf{3DRES}} \\
		\cmidrule(r){2-3} \cmidrule(l){4-6}
		& \textbf{0.25} & \textbf{0.50} & 
		\textbf{0.25} & \textbf{0.50} &
		\textbf{mIoU} \\ 
		\midrule
		$F^{(4)}_{\mathrm{v}}$ & 56.94 & 45.98 & 58.87 & 50.97 & 44.94 \\
		$\mathcal{K}^{\mathrm{t}2\mathrm{v}}_{\lambda}$ & 58.06 & 46.92 & 60.25 & 51.84 & 46.13 \\
		\midrule
		$F_{\mathrm{t}}$ & 57.93 & 46.71 & 60.02 & 52.42 & 46.27 \\
		$\mathcal{K}^{\mathrm{v}2\mathrm{t}}_{\lambda}$ & \textbf{58.47} & \textbf{47.89} & \textbf{60.41} & \textbf{52.52} & \textbf{46.39} \\
		
		\bottomrule
	\end{tabular}
	
	\caption{Comparison of localization accuracy when integrating different feature configurations from PLDA into CLDA.}
	\label{as_max}
\end{table}

As shown in Table~\ref{as_max}, both $F^{(4)}_{\mathrm{v}}$ and $F_{\mathrm{t}}$, when processed by DiffAtten, significantly improve localization accuracy over their original counterparts, validating the effectiveness of cross-modal differential attention. Notably, text-based $\mathcal{K}^{\mathrm{v}2\mathrm{t}}_{\lambda}$ achieves optimal performance through two mechanisms: 
(i) $F_{\mathrm{t}}$ provides more precise target information; 
(ii) PLDA enhances discriminative localization cues in text while suppressing irrelevant features and improves feature discriminability.

\begin{table}[htbp]
	\centering
	
	\begin{tabular}{@{\extracolsep{\fill}} 
			>{\centering\arraybackslash}p{0.23\linewidth}
			>{\centering\arraybackslash}p{0.09\linewidth}
			*{3}{>{\centering\arraybackslash}p{0.09\linewidth}} 
			>{\centering\arraybackslash}p{0.09\linewidth}
		}
		\toprule 
		\textbf{Component} &  
		\multicolumn{2}{c}{\textbf{3DREC}} & 
		\multicolumn{3}{c}{\textbf{3DRES}} \\
		\cmidrule(r){2-3} \cmidrule(l){4-6}
		& \textbf{0.25} & \textbf{0.50} & 
		\textbf{0.25} & \textbf{0.50} &
		\textbf{mIoU} \\ 
		\midrule
		
		$\star$  & 56.64 & 45.81 & 58.54 & 50.63 & 44.85 \\
		PLDA+CLDA  & \textbf{58.47} & \textbf{47.89} & \textbf{60.41} & \textbf{52.52} & \textbf{46.39} \\	
		
		\bottomrule
	\end{tabular}
	\caption{Performance comparison of our dual-level differential attention and standard cross-attention ($\star$: standard Transformer-based \cite{NIPS2017_3f5ee243}).}
	\label{plda_add_clda}
\end{table}

Table~\ref{plda_add_clda} shows that replacing the proposed dual-level differential attention with standard Transformer attention~\cite{NIPS2017_3f5ee243} degrades Overall@0.50 by 2.08\% and 1.89\% on 3DREC and 3DRES, respectively. This degradation reveals that standard attention fails to preserve implicit semantic cues and suppress irrelevant spatial interference. In contrast, our framework explicitly addresses these issues by leveraging PLDA for extracting implicit localization cues and CLDA for dynamic spatial filtering, validating the need for a dual-level design.

\begin{figure}
	\centering 
	\includegraphics[width=\linewidth]{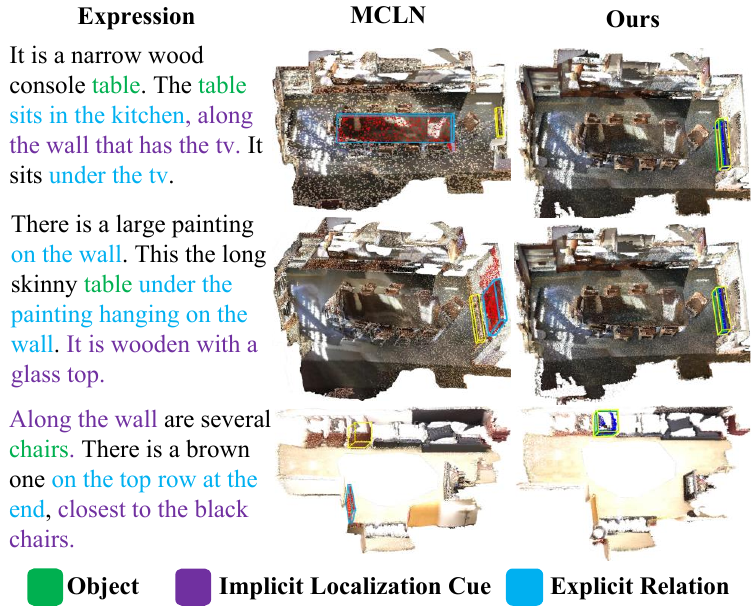}
	\caption{Comparison of PC-CrossDiff (ours) and MCLN (baseline) for 3DREC and 3DRES with implicit cues in complex multi-object scenes. GT boxes are yellow.}
	\label{pcd_net_visu}
\end{figure}

\section{Qualitative Results}

We compare PC-CrossDiff with the state-of-the-art dual-task method MCLN~\cite{qianMultibranchCollaborativeLearning2024} on complex multi-object scenes. As shown in Fig.~\ref{pcd_net_visu}, MCLN mislocalizes targets due to missing implicit cues (Rows 1–2). In contrast, PC-CrossDiff successfully captures target locations, particularly for implicit or occluded instances (Rows 1–3). The superior performance stems from two key mechanisms: bidirectional cross-modal interaction in PLDA mitigates semantic ambiguity in implicit localization, while dynamic spatial weighting in CLDA resolves spatial relationship ambiguity. These components jointly enable robust and accurate localization in complex multi-object scenes.

\section{Conclusion}

In this paper, we propose PC-CrossDiff, a novel dual-hierarchy optimization framework, to address the challenges of inadequate implicit semantic parsing and spatial interference in 3DVG for multi-object scenes. PLDA introduces a text-point cloud bidirectional differential attention mechanism to adaptively extract implicit localization cues, mitigating semantic ambiguity caused by explicit relation parsing in conventional approaches. CLDA enhances localization robustness through dynamic spatial relationship modeling with suppression of irrelevant interference, enabling efficient reasoning in multi-object scenes. A unified multi-task model integrating 3DREC and 3DRES achieves state-of-the-art performance on ScanRefer and NR3D/SR3D benchmarks, demonstrating its effectiveness in handling multi-object scenes.

\section{Acknowledgements}

This work was supported in part by the National Natural Science Foundation of China under Grant 62176224, Grant 62176092, Grant 62222602, Grant 62306165, and Grant 62172234; in part by the Fundamental Research Funds for the Central Universities under Grant 20720250031; in part by the Science and Technology on Sonar Laboratory under Grant 2024-JCJQ-LB-32/07; in part by the Guizhou Provincial Department of Science and Technology under Grant Qian Ke He QN [2025] 337; and in part by the Tongren Municipal Science and Technology Bureau under Grant Tongshi [2025] 54.

\bibliography{aaai2026}

\clearpage
\appendix
\section{Appendix}

\subsection{1~~Methodological Supplementary Explanation}

\paragraph{Text-to-Visual Cross-modal Differential Attention.} 
This component enhances point-level visual features by using textual features $T \in \mathbb{R}^{l \times d_\mathrm{sem}}$ as queries ($Q$) and high-level visual features $F_\mathrm{v}^{(4)} \in \mathbb{R}^{n_\mathrm{sem} \times d_\mathrm{sem}}$ as keys ($K$) and values ($V$). Specifically, we project $T$ and $F_\mathrm{v}^{(4)}$ via learnable matrices $W^Q, W^K, W^V \in \mathbb{R}^{d_\mathrm{sem} \times d_\mathrm{sem}}$:
\begin{equation}
	\begin{aligned}
		Q &= T W^Q, & 
		K &= F_\mathrm{v}^{(4)} W^K, & 
		V &= F_\mathrm{v}^{(4)} W^V.
	\end{aligned}
	\label{eq:projection2}
\end{equation}

The projected tensors are reshaped into a multi-head structure compatible with dual-channel differential attention (head dimension $d_\mathrm{h} = d_\mathrm{sem}/(2N_\mathrm{h})$, $N_\mathrm{h} = 8$):
\begin{equation}
	\begin{aligned}
		Q &\triangleq \text{reshape}\left(Q, [l, 2N_\mathrm{h}, d_\mathrm{h}]\right)^\intercal = [Q_1, Q_2] \\
		K &\triangleq \text{reshape}\left(K, [n_\mathrm{sem}, 2N_\mathrm{h}, d_\mathrm{h}]\right)^\intercal = [K_1, K_2] \\
		V &\triangleq \text{reshape}\left(V, [n_\mathrm{sem}, N_\mathrm{h}, 2d_\mathrm{h}]\right)^\intercal.
	\end{aligned}
	\label{eq:q_k_v2}
\end{equation}

We compute two attention kernels as follows:
\begin{equation*}
	\begin{aligned}
		A_1=\text{softmax}\left(\frac{Q_1 K_1^\intercal}{\sqrt{d_\mathrm{h}}}\right), 
		A_2=\text{softmax}\left(\frac{Q_2 K_2^\intercal}{\sqrt{d_\mathrm{h}}}\right).
	\end{aligned}
\end{equation*}

To adaptively regulate feature filtering during training, we introduce parameterized differential coefficients:
\begin{equation}
	\text{DiffAttn}(T, F_{\text{v}}^{(4)}) = (A_1 - \lambda \odot A_2)V,
	\label{eq:DiffAttn_t_v2}
\end{equation}
where $\lambda \in \mathbb{R}^{l \times n_\mathrm{sem} \times N_\mathrm{h}}$ is computed as:
\begin{equation*}
	\lambda = \exp\left(\sum (\lambda_{q1} \odot \lambda_{k1})\right) - \exp\left(\sum (\lambda_{q2} \odot \lambda_{k2})\right),
\end{equation*}
with $\lambda_{q1}, \lambda_{k1}, \lambda_{q2}, \lambda_{k2} \in \mathbb{R}^{d_\mathrm{h}}$ being learnable vectors that modulate the differential strength. 
This mechanism enables the model to suppress irrelevant visual features guided by the textual features, thereby enhancing the discriminability of local visual representations.

The output of Eq.~\eqref{eq:DiffAttn_t_v2} is processed through layer normalization, dimension transformation, and linear projection to yield the enhanced point-level visual features $\mathcal{K}_\mathrm{t2v} \in \mathbb{R}^{l \times d_\mathrm{sem}}$.

\paragraph{Implementation Mechanism Analysis of CLDA.}
As shown in Algorithm~\ref{alg:clda}, after cross-modal alignment, the spatial coordinates of visual features $F'_{\mathrm{v}} \in \mathbb{R}^{n_\mathrm{p} \times d_\mathrm{mod}}$ are processed by Spatial Relation Modeling (SRM) to extract two key components: inter-cluster spatial relation features $F'_{\mathrm{oRel}} \in \mathbb{R}^{N_{\mathrm{clust}} \times d_{\mathrm{mod}}}$ and candidate object region features $F'_{\mathrm{ctr}} \in \mathbb{R}^{N_{\mathrm{clust}} \times d_{\mathrm{mod}}}$.

\begin{algorithm}[H]
	\caption{The Working Process of CLDA.}
	\label{alg:clda}
	\begin{algorithmic}[1]
		\STATE \textbf{Step 1: Spatial Relation Modeling (SRM)}
		\STATE \quad Compute cluster centroids $\mathbf{O} \in \mathbb{R}^{B \times N_{\mathrm{clust}} \times 3}$ via FPS~\cite{qiPointNetDeepHierarchical2017}.
		\STATE \quad Partition points into $N_{\mathrm{clust}}$ clusters of size $S_\mathrm{clust} = n_\mathrm{p}/N_{\mathrm{clust}}$ ($n_\mathrm{p} = 1024$).
		\STATE \quad Extract intra-cluster relations $F_{\mathrm{iRel}}$ using encoder.
		\STATE \quad Compute inter-cluster spatial relationships:
		\[
		F^{\prime}_{\mathrm{oRel}} = \mathrm{DGCNN}(F_{\mathrm{iRel}}, \mathbf{O}) \in \mathbb{R}^{B \times N_{\mathrm{clust}} \times d_{\mathrm{mod}}}.
		\]
		\STATE \quad Compute the candidate object region features:
		\[
		F^{\prime}_{\mathrm{ctr}} = \mathbf{O} \mathbf{W}_\mathrm{o} + \mathbf{b}_\mathrm{o} \in \mathbb{R}^{B \times N_{\mathrm{clust}} \times d_{\mathrm{mod}}},
		\]
		\quad where $\mathbf{W}_\mathrm{o} \in \mathbb{R}^{3 \times d_{\mathrm{mod}}}$.
		
		\STATE \textbf{Step 2: Dual-Branch LDA for Dynamic Filtering and Enhancement}
		\STATE \quad \textit{\% Branch 1: Spatial Filtering.}
		
		\STATE \quad $Q \leftarrow F^{\prime}_{\mathrm{v}}$, \quad $K, V \leftarrow F^{\prime}_{\mathrm{oRel}}$
		\STATE \quad Compute suppression factor:
		\[
		\lambda = \exp\left(\sum (\lambda_{q1} \odot \lambda_{k1})\right) - \exp\left(\sum (\lambda_{q2} \odot \lambda_{k2})\right),
		\]
		\quad with $\lambda_{q1}, \lambda_{k1}, \lambda_{q2}, \lambda_{k2} \in \mathbb{R}^{d_{\mathrm{h}}}$ learnable.
		\STATE \quad Compute attention maps and differential output:
		\[
		A_1 = \mathrm{softmax}\left(\frac{Q K^\intercal}{\sqrt{d_{\mathrm{h}}}}\right), \quad
		A_2 = \mathrm{softmax}\left(\frac{Q K^\intercal}{\sqrt{d_{\mathrm{h}}}}\right),
		\]
		\[
		F^{\prime}_{\mathrm{oRel}}{}^{\text{filt}} \leftarrow (A_1 - \lambda \odot A_2)V.
		\]
		\STATE \quad Update: $\tilde{F}^{\prime}_{\mathrm{v}} \leftarrow F^{\prime}_{\mathrm{v}} + \mathrm{MLP}(F^{\prime}_{\mathrm{oRel}}{}^{\text{filt}})$
		
		\STATE \quad \textit{\% Branch 2: Target Enhancement.}
		\STATE \quad $Q \leftarrow F^{\prime}_{\mathrm{t}}$, \quad $K, V \leftarrow F^{\prime}_{\mathrm{ctr}}$
		\STATE \quad Compute:
		\[
		A_1 = \mathrm{softmax}\left(\frac{Q K^\intercal}{\sqrt{d_{\mathrm{h}}}}\right), \quad
		A_2 = \mathrm{softmax}\left(\frac{Q K^\intercal}{\sqrt{d_{\mathrm{h}}}}\right), 
		\]
		\[
		F^{\prime}_{\mathrm{ctr}}{}^{\text{enh}} \leftarrow (A_1 - \lambda \odot A_2)V.
		\]
		\STATE \quad Update: $\tilde{F}^{\prime}_{\mathrm{t}} \leftarrow F^{\prime}_{\mathrm{t}} + F^{\prime}_{\mathrm{ctr}}{}^{\text{enh}}$
		
		\RETURN $\tilde{F}^{\prime}_{\mathrm{v}}, \tilde{F}^{\prime}_{\mathrm{t}}$
	\end{algorithmic}
\end{algorithm}

The Cluster-Level Differential Attention (CLDA) module adopts a dual-branch Localization-aware Differential Attention (LDA) mechanism to jointly refine spatial and semantic representations. In the first stage, \textit{spatial filtering}, aligned visual features $F'_{\mathrm{v}}$ serve as queries ($Q$), while $F'_{\mathrm{oRel}}$ acts as both keys ($K$) and values ($V$). The differential attention operation is formulated as:
\begin{equation}
	\text{DiffAttn}(F'_{\mathrm{v}}, F'_{\mathrm{oRel}}) = (A_1 - \lambda \odot A_2)V,
	\label{eq:f_v_oRel2}
\end{equation}
where $A_1$ and $A_2$ denote the attention maps computed from $F'_{\mathrm{v}}$ and $F'_{\mathrm{oRel}}$, respectively, and $\lambda \in \mathbb{R}^{N_{\mathrm{clust}} \times N_{\mathrm{clust}}}$ is a learnable parameter optimized during training. This mechanism suppresses irrelevant spatial relations through adaptive learning and modulation of $\lambda$, enabling selective suppression of irrelevant localization cues and preserving location-sensitive features. The output is passed through a multi-layer perceptron (MLP), added to $F'_{\mathrm{v}}$, and further refined via multi-head self-attention to produce spatially enhanced visual features $\tilde{F}_{\mathrm{v}}^{\prime}$.

In the second stage, \textit{target enhancement}, textual features $F'_{\mathrm{t}}$ are used as queries ($Q$), while $F'_{\mathrm{ctr}}$ serves as keys ($K$) and values ($V$). The same differential attention mechanism is applied:
\begin{equation}
	\text{DiffAttn}(F'_{\mathrm{t}}, F'_{\mathrm{ctr}}) = (A_1 - \lambda \odot A_2)V,
\end{equation}
which selectively extracts discriminative region features from $F'_{\mathrm{ctr}}$ and injects them into $F'_{\mathrm{t}}$. These enriched features are then refined via multi-head self-attention to yield enhanced textual features $\tilde{F}_{\mathrm{t}}^{\prime}$.

Finally, both modalities undergo additional refinement through transformer blocks and layer normalization. This design leverages SRM to capture global spatial configurations and candidate object regions, which are selectively integrated into aligned visual and textual features via the LDA filtering mechanism. During decoding, the enhanced features enable precise parsing of object identities and spatial relationships, significantly reducing interference from irrelevant localization features and improving performance in complex multi-object scenes.

\paragraph{Mechanism of Noise Suppression in Cross-modal DiffAttn.} 
Cross-modal differential attention (DiffAttn) is designed to suppress irrelevant visual noise during 3D object localization by leveraging a dual-channel decomposition mechanism that explicitly models the cancellation of distracting attention scores, inspired by the noise-canceling property of differential attention~\cite{yeDifferentialTransformer2024}.
Unlike standard self-attention, which aggregates all context signals uniformly, DiffAttn separates the query and key-value representations into two distinct subspaces, $Q_1, Q_2$ and $K_1, K_2$, using a structured reshaping operation (Eq.~\ref{eq:q_k_v2}). This enables the computation of two independent attention matrices:
\begin{equation*}
	A_1 = \text{softmax}\left(\frac{Q_1 K_1^\intercal}{\sqrt{d_\mathrm{h}}}\right), 
	A_2 = \text{softmax}\left(\frac{Q_2 K_2^\intercal}{\sqrt{d_\mathrm{h}}}\right),
\end{equation*}
where $A_1$ captures semantic alignment between textual descriptions and visual features, while $A_2$ encodes potential distractive or inconsistent responses.

The critical innovation lies in the differential suppression mechanism:
\begin{equation}
	\text{DiffAttn}(T, F_{\text{v}}^{(4)}) = (A_1 - \lambda \odot A_2)V,
\end{equation}
where the learnable parameter $\lambda \in \mathbb{R}^{d_\mathrm{h}}$ dynamically regulates the strength of $A_2$'s suppression.
The trainable parameter $\lambda$ is computed as:
\[
\lambda = \exp\left(\sum (\lambda_{q1} \odot \lambda_{k1})\right) - \exp\left(\sum (\lambda_{q2} \odot \lambda_{k2})\right).
\]
During training, both $A_1$ and $\lambda \odot A_2$ contain components corresponding to spatially widespread, task-irrelevant noise (e.g.,
background clutter or features unrelated to object localization). The differential subtraction effectively suppresses the shared, broadly
distributed noise in $A_1$, thereby reducing irrelevant interference. This differential suppression mechanism has proven effective in the
Differential Transformer~\cite{yeDifferentialTransformer2024}. By learning $\lambda$ end to end, the model adaptively controls the suppression
strength of distracting attention patterns, promoting improved localization accuracy. Consequently, the model learns to optimally suppress
features unrelated to object localization, preserving discriminative information and enhancing the representational quality of the filtered
features, thereby improving their discriminability in multi-object scenes.

\subsection{2~~Extended Performance Comparisons}

\paragraph{3DREC and 3DRES on SR3D.}
As shown in Table~\ref{SR3D}, PC-CrossDiff achieves state-of-the-art performance on both the 3DREC and 3DRES tasks on the SR3D dataset. It
attains $61.27\%$ on 3DREC and $62.04\%$ on 3DRES at Overall@0.50, representing absolute improvements of $3.97$ and $4.16$ over the previous
SOTA method, MCLN~\cite{qianMultibranchCollaborativeLearning2024}, respectively. These improvements highlight the model's strong
generalization ability in jointly modeling object recognition and spatial localization.

\begin{table}[!h]
	\centering
	
	\begin{tabular}{@{\extracolsep{\fill}} 
			>{\centering\arraybackslash}p{0.23\linewidth}
			*{4}{>{\centering\arraybackslash}p{0.09\linewidth}} 
			>{\centering\arraybackslash}p{0.1\linewidth}
		}
		
		\toprule 
		\textbf{Method}  &  
		\multicolumn{2}{c}{\textbf{3DREC}} & 
		\multicolumn{3}{c}{\textbf{3DRES}}  \\
		
		\cmidrule(lr){2-3} \cmidrule(lr){4-6} &    
		
		\textbf{0.25} & \textbf{0.50} & 
		\textbf{0.25} & \textbf{0.50} &
		\multicolumn{1}{c}{\textbf{mIoU}}  \\ 
		\midrule
		
		MCLN & 68.43 & 57.30 & 64.70 & 57.88 & 49.84 \\
		Ours  & \textbf{70.95} & \textbf{61.27} & \textbf{68.70} & \textbf{62.04} & \textbf{53.43} \\
		\bottomrule
	\end{tabular}
	\caption{Performance comparison of 3DREC and 3DRES on the SR3D dataset.}
	
	\label{SR3D}
\end{table}

\paragraph{3DREC and 3DRES on NR3D.}
We further evaluate PC-CrossDiff compared to baseline methods on the NR3D dataset for both 3DREC and 3DRES. 

\begin{table}[!h]
	\centering
	
	\begin{tabular}{@{\extracolsep{\fill}} 
			>{\centering\arraybackslash}p{0.23\linewidth}
			*{4}{>{\centering\arraybackslash}p{0.09\linewidth}} 
			>{\centering\arraybackslash}p{0.1\linewidth}
		}
		
		\toprule 
		\textbf{Method}  &  
		\multicolumn{2}{c}{\textbf{3DREC}} & 
		\multicolumn{3}{c}{\textbf{3DRES}}  \\
		
		\cmidrule(lr){2-3} \cmidrule(lr){4-6} &    
		
		\textbf{0.25} & \textbf{0.50} & 
		\textbf{0.25} & \textbf{0.50} &
		\multicolumn{1}{c}{\textbf{mIoU}}  \\ 
		\midrule
		
		MCLN$^\dag$  & 53.75 & 45.82 & 51.79 & 46.58 & 40.47 \\
		
		Ours & \textbf{59.91} & \textbf{50.48} & \textbf{57.56} & \textbf{52.64} & \textbf{46.00} \\
		
		\bottomrule
	\end{tabular}
	\caption{Performance comparison of joint 3DREC and 3DRES tasks on the NR3D dataset ($^\dag$reproduced results; 200 epochs).}
	\label{NR3D_res_rec}
\end{table}

As shown in Table~\ref{NR3D_res_rec}, our method achieves state-of-the-art performance on both 3DREC and 3DRES, with absolute improvements of
$4.66$ and $6.06$ over MCLN~\cite{qianMultibranchCollaborativeLearning2024} on the Overall@0.50 metric, respectively. These gains highlight
the model's strong generalization ability.

\subsection{3~~Extended Ablation Studies}

\subsection{3.1~~Multi-object Scenes Localization Ablation}

To evaluate the impact of PLDA and CLDA on localization performance in complex multi-object scenes, we conduct ablation studies on the \textit{Multiple} subset of ScanRefer for 3DREC and 3DRES.

\begin{table}[htbp]
	\centering
	
	\begin{tabular}{@{\extracolsep{\fill}}
			>{\centering\arraybackslash}p{0.12\linewidth}
			>{\centering\arraybackslash}p{0.12\linewidth}
			*{2}{>{\centering\arraybackslash}p{0.11\linewidth}}
			*{2}{>{\centering\arraybackslash}p{0.11\linewidth}}
		}
		\toprule 
		\multicolumn{2}{c}{\textbf{PLDA / CLDA}} & 
		\multicolumn{2}{c}{\textbf{3DREC}} & 
		\multicolumn{2}{c}{\textbf{3DRES}} \\
		\cmidrule(r){1-2} \cmidrule(lr){3-4} \cmidrule(l){5-6}
		& & \textbf{0.25} & \textbf{0.50} & \textbf{0.25} & \textbf{0.50} \\
		\midrule
		w/o & w/o & 51.96 & 40.76 & 53.28 & 45.88 \\
		$\checkmark$ & w/o & 52.35 & 41.51 & 53.84 & 46.67 \\
		w/o & $\checkmark$ & 52.74 & 41.50 & 54.57 & 47.56 \\
		$\checkmark$ & $\checkmark$ & \textbf{53.59} & \textbf{43.12} & \textbf{55.48} & \textbf{47.98} \\
		\bottomrule
	\end{tabular}
	\caption{Ablation study of PLDA and CLDA on ScanRefer-Multiple for 3DREC and 3DRES. $\checkmark$: included; \text{w/o}: no CLDA.}
	
	\label{ScanRefer_Imp_Mul}
\end{table}

As shown in Table~\ref{ScanRefer_Imp_Mul}, introducing PLDA alone improves 3DREC@0.25 from 51.96\% to 52.35\% (+0.39\%) and 3DRES@0.25 from 53.28\% to 53.84\% (+0.56\%), demonstrating its effectiveness in leveraging textual semantic cues for localization in multi-object scenes.

When only CLDA is used, 3DREC@0.25 increases to 52.74\% (+0.78\%), 3DREC@0.50 to 41.50\% (+0.74\%), and 3DRES@0.25 to 54.57\% (+1.29\%), indicating that CLDA’s spatial relation filtering effectively reduces semantic ambiguity and enhances understanding of 3D spatial relationships.

Jointly enabling PLDA and CLDA yields the best performance: 53.59\% (3DREC@0.25), 43.12\% (3DREC@0.50), 55.48\% (3DRES@0.25), and 47.98\% (3DRES@0.50), significantly outperforming either module alone. This confirms the efficacy of the point-cluster dual-level collaborative design: PLDA disentangles implicit semantic clues from text, while CLDA refines cluster-level spatial relations and object features, together improving robustness and accuracy in complex multi-object scenes.

\subsection{3.2~~Robustness of CLDA to Cluster Size Variations}

To investigate the impact of cluster size (i.e., cluster count $N_{\mathrm{clust}} = 1024 / S_{\mathrm{clust}}$) on CLDA performance, we conduct ablation studies by varying $N_{\mathrm{clust}}$ from 32 to 256, corresponding to cluster sizes of $S_{\mathrm{clust}} = 32$ and
$S_{\mathrm{clust}} = 4$, respectively. As shown in Table~\ref{as_group_num2}, CLDA remains effective across different cluster configurations
and improves the 0.50 metrics and mIoU over the baseline without CLDA for all tested settings.

\begin{table}[htbp]
	\centering
	
	\begin{tabular}{@{\extracolsep{\fill}}
			>{\centering\arraybackslash}p{0.1\linewidth}
			>{\centering\arraybackslash}p{0.1\linewidth}
			*{2}{>{\centering\arraybackslash}p{0.09\linewidth}}
			*{3}{>{\centering\arraybackslash}p{0.09\linewidth}}
		}
		\toprule 
		\multicolumn{2}{c}{\textbf{CLDA}} & 
		\multicolumn{2}{c}{\textbf{3DREC}} & 
		\multicolumn{3}{c}{\textbf{3DRES}} \\
		\cmidrule(r){1-2} \cmidrule(lr){3-4} \cmidrule(l){5-7}
		\textbf{$N_{\mathrm{clust}}$} & 
		\textbf{$S_{\mathrm{clust}}$} & 
		\textbf{0.25} & \textbf{0.50} & 
		\textbf{0.25} & \textbf{0.50} & 
		\textbf{mIoU} \\
		\midrule
		/ & / & 56.76 & 44.86 & 58.43 & 49.66 & 43.85 \\
		256 & 4 & 56.11 & 46.37 & 57.83 & 50.88 & 44.66 \\
		128 & 8 & 58.15 & 47.09 & 60.21 & 52.40 & 46.16 \\
		64 & 16 & \textbf{58.47} & \textbf{47.89} & \textbf{60.41} & \textbf{52.52} & \textbf{46.39} \\
		32 & 32 & 57.36 & 47.01 & 59.25 & 51.53 & 45.54 \\
		\bottomrule
	\end{tabular}
	\caption{Ablation study of cluster configurations ($N_{\mathrm{clust}}$ and $S_{\mathrm{clust}}$) on 3DREC and 3DRES tasks. \text{/} denotes no CLDA.}
	\label{as_group_num2}
\end{table}

When $N_{\mathrm{clust}} = 256$ ($S_{\mathrm{clust}} = 4$), the fine-grained partitioning may introduce noisy, low-level spatial relations.
Although 3DREC@0.25 decreases to 56.11\% (vs. 56.76\% in the baseline), CLDA still improves 3DREC@0.50 from 44.86\% to 46.37\%, 3DRES@0.50
from 49.66\% to 50.88\%, and mIoU from 43.85\% to 44.66\%, suggesting robustness under over-segmentation.

Conversely, with $N_{\mathrm{clust}} = 32$ ($S_{\mathrm{clust}} = 32$), coarse clustering may result in loss of fine-grained geometric details. Nonetheless, CLDA improves 3DREC@0.25 to 57.36\% (+0.60\%) and 3DRES@0.25 to 59.25\% (+0.82\%), demonstrating its capacity to recover useful spatial relationships despite information loss.

Further analysis reveals minimal variation across intermediate settings: as $N_{\mathrm{clust}}$ decreases from 128 to 64 ($S_{\mathrm{clust}}$ increases from 8 to 16), 3DREC@0.25 increases by 0.32\% (from 58.15\% to 58.47\%), and 3DREC@0.50 rises by 0.80\%. This underscores CLDA’s high stability under varying clustering configurations.

In summary, CLDA maintains consistent performance across diverse cluster sizes, demonstrating robustness under both fine-grained and coarse partitioning. Its stable behavior arises from the ``filter-then-enhance'' mechanism, enabling reliable spatial relation modeling in complex multi-object scenes.

\begin{algorithm}[h]
	\caption{Construction of the Implicit Subset from ScanRefer.}
	\label{alg:implicit_subset}
	\begin{algorithmic}[1]
		\REQUIRE 
		$\mathcal{D} = \{(s_i, o_i, n_i, d_i)\}_{i=1}^N$: Scene-object-description triplets, \\
		$P_{\text{explicit}}$: Explicit spatial markers (case-insensitive regex), \\
		$\mathcal{P}_{\text{implicit}}$: Pre-compiled patterns for implicit relations, \\
		$\text{Qwen}$: Interface to large language model.
		\ENSURE 
		$\mathcal{S}_{\text{implicit}} \subseteq \mathcal{D}$: Descriptions with implicit spatial cues.
		
		\STATE Compile $P_{\text{explicit}}$ and $\mathcal{P}_{\text{implicit}}$ into case-insensitive regular expressions.
		\STATE Initialize $\mathcal{S}_{\text{implicit}} \leftarrow \emptyset$
		
		\FOR{each $d_i \in \mathcal{D}$}
		\IF{$d_i$ contains any $p \in P_{\text{explicit}}$}
		\STATE \textbf{continue}
		\ENDIF
		\STATE $r_{\text{rule}} \leftarrow \texttt{match\_regex}(d_i, \mathcal{P}_{\text{implicit}})$
		\IF{use\_qwen and $\neg r_{\text{rule}}$}
		\STATE $r_{\text{final}} \leftarrow \texttt{parse}(\texttt{QwenAPI}(\texttt{build\_prompt}(d_i)))$
		\ELSE
		\STATE $r_{\text{final}} \leftarrow r_{\text{rule}}$
		\ENDIF
		\IF{$r_{\text{final}}$}
		\STATE $\mathcal{S}_{\text{implicit}} \leftarrow \mathcal{S}_{\text{implicit}} \cup \{ (s_i, o_i, n_i, d_i) \}$
		\ENDIF
		\ENDFOR
		\RETURN $\mathcal{S}_{\text{implicit}}$
	\end{algorithmic}
\end{algorithm}

\subsection{3.3~~Construction of the Implicit Subset from ScanRefer}

To rigorously evaluate the 3DVG model's ability to interpret implicit spatial relationships in natural language, we construct an implicit subset from ScanRefer via a hybrid rule-based and LLM-assisted detection pipeline. The goal is to identify descriptions expressing spatial relationships via physical interactions (e.g., \textit{resting on}, \textit{attached to}), functional dependencies (e.g., \textit{supporting}, \textit{holding}), or contextual associations (e.g., \textit{embedded in}), excluding explicit spatial terms like \textit{above}, \textit{left}, or \textit{between}. 

The detection process, formalized in Algorithm~\ref{alg:implicit_subset}, starts by compiling two complementary pattern libraries. $P_{\text{explicit}}$ is converted into case-insensitive regular expressions for efficient filtering of explicit cues. $\mathcal{P}_{\text{implicit}}$ covers three semantic categories of implicit relations and is pre-compiled for fast matching.

All descriptions containing any explicit marker are immediately excluded. For the remaining candidates, we apply a rule-first, LLM-backup strategy: if no rule match is found, the description is sent to the Qwen-turbo\cite{ahmed2025qwen} API via a structured prompt to determine whether it expresses an implicit spatial relationship, with output parsed into binary labels. Conflicts between rule and LLM predictions are flagged for manual review to ensure labeling accuracy.

The final implicit subset contains 2,490 instances, including 1,989 from the original training split and 501 from the original validation split. In our experimental setup, only the 501 instances from the original ScanRefer validation set are used as the held-out test set for evaluating implicit spatial reasoning. Critically, this subset is never used during training, ensuring unbiased evaluation of the model’s generalization to unseen implicit descriptions. This design ensures that performance gains on implicit cues reflect genuine understanding rather than memorization, improving the reliability of the evaluation.

\end{document}